\documentclass[journal]{IEEEtran}
\usepackage{flushend}
\usepackage{ifpdf}
\usepackage{graphicx}
\ifpdf
 \graphicspath{{Figs/}{Figs/PDF/}}
\else
 \graphicspath{{Figs/}}
\fi

\usepackage{hhline}

\usepackage{calc}
\usepackage{enumitem}

\usepackage{hyperref}
\hypersetup{
	colorlinks = true,
	linkcolor = {blue},
	citecolor = {blue}
}
\usepackage{url}

\usepackage{nomencl}
\usepackage{multirow}

\usepackage{subcaption}
\usepackage{todonotes}

\usepackage{array}
\usepackage{booktabs}

\newcolumntype{P}[1]{>{\centering\arraybackslash}p{#1}}

\usepackage[flushleft]{threeparttable}
\bstctlcite{IEEEexample:BSTcontrol}
\usepackage{cite}

\ifCLASSINFOpdf
\else
\fi
\begin{document}
%
\title{Robotic Tactile Perception of Object Properties: A Review}
%
%
%

\author{Shan~Luo\textsuperscript{*}, Joao Bimbo,
	Ravinder~Dahiya
	and~Hongbin~Liu
\thanks{\textsuperscript{*} Corresponding author.}
\thanks{S. Luo and H. Liu are with the Centre for Robotics Research, Department of Informatics, King's College London, London WC2R 2LS, U.K. (e-mail: shan.luo, hongbin.liu@kcl.ac.uk). S. Luo is also with School of Civil Engineering and School of Computing, University of Leeds, Leeds LS2 9JT, U.K., and Brigham and Women's Hospital, Harvard Medical School,  75 Francis St, Boston, MA 02115.}
\thanks{J. Bimbo is with the Department of Advanced Robotics, Istituto Italiano di Tecnologia, 16163 Genova, Italy (e-mail: joao.bimbo@iit.it).} 
\thanks{R. S. Dahiya is with the Electronics and Nanoscale Engineering Research Division, University of Glasgow, Glasgow G12 8QQ, U.K. (e-mail: ravinder.dahiya@glasgow.ac.uk).
}
}

%
%

\markboth{Preprint submitted to Elsevier}%
{Shell \MakeLowercase{\textit{et al.}}: Bare Demo of IEEEtran.cls for IEEE Journals}
%



\maketitle

\begin{abstract}
Touch sensing can help robots understand their surrounding environment, and in particular the objects they interact with. To this end, roboticists have, in the last few decades, developed several tactile sensing solutions, extensively reported in the literature. Research into interpreting the conveyed tactile information has also started to attract increasing attention in recent years. However, a comprehensive study on this topic is yet to be reported. In an effort to collect and summarize the major scientific achievements in the area, this survey extensively reviews current trends in robot tactile perception of object properties.
Available tactile sensing technologies are briefly presented before an extensive review on tactile recognition of object properties.
The object properties that are targeted by this review are shape, surface material and object pose.
The role of touch sensing in combination with other sensing sources is also discussed. 
In this review, open issues are identified and future directions for applying tactile sensing in different tasks are suggested.
\end{abstract}

\begin{IEEEkeywords}
Tactile sensing, robot tactile systems, object recognition, sensor fusion, survey.
\end{IEEEkeywords}

%
\IEEEpeerreviewmaketitle

\section{Introduction}
\label{sec:intro}
%
%
%
%
\IEEEPARstart{T}{he} sense of touch is an irreplaceable source of information for humans while exploring the environment in their close vicinity. It conveys diverse sensory information, such as pressure, vibration, pain and temperature, to the central nervous system, assisting humans in perceiving their surroundings and avoiding potential injuries \cite{dahiya2012robotic}. Research has shown that, compared to vision and audition, the human sense of touch is superior at processing material characteristics and detailed shapes of objects \cite{lederman2009haptic,dahiya2012robotic}. As for humans, it is essential that robots are also equipped with advanced touch sensing in order to be aware of their surroundings, keep away from potentially destructive effects and provide information for subsequent tasks such as in-hand manipulation.

\begin{figure}
	\includegraphics[width=.6\textwidth]{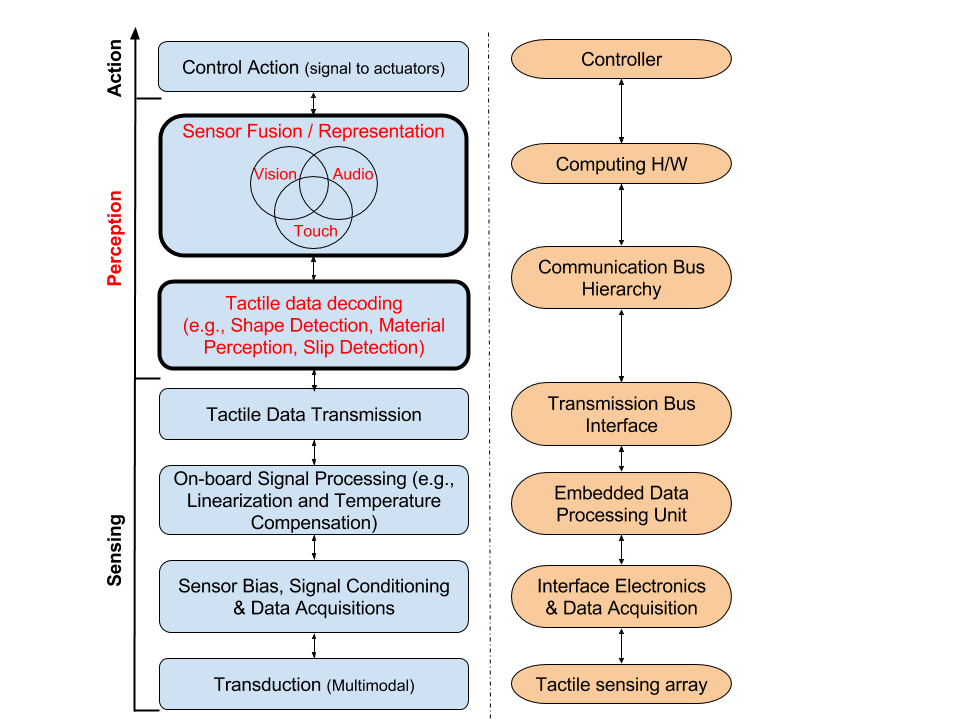}
	\caption{Hierarchical functional (left) and structural (right) block diagrams of robotic tactile sensing system \cite{dahiya2013directions,dahiya2010tactile}. The perception blocks have been highlighted that the paper is contributing.}
	\label{fig:tactilesensingsystem}
\end{figure}

A general block diagram of a tactile sensing system \cite{dahiya2013directions,dahiya2010tactile} is illustrated in Fig.~\ref{fig:tactilesensingsystem}.
A tactile sensing system here refers to a system where a robot uses tactile sensors to sense the ambient stimuli through touch, acquiring information on the properties of objects, such as shape and material, providing action related information, such as object localization and slippage detection.
On the left side of Fig.~\ref{fig:tactilesensingsystem}, the tactile sensing process is divided into functional blocks that depict sensing, perception and action at different levels. We follow existing literature \cite{lederman2009haptic}, using the term ``perception'' to refer to the process of observing object properties through sensing. The right side of Fig.~\ref{fig:tactilesensingsystem} shows the structural blocks of hardware that correspond to those functional blocks. The sensing process transduces the external stimuli (e.g., pressure, vibration and thermal stimulus), into changes on the sensing elements of the tactile sensors \cite{dahiya2013directions,dahiya2010tactile}. This data is acquired, conditioned and processed using an embedded data processing unit, and then transferred to the higher perception level to construct a world model, perceive the properties of interacted objects, (e.g., shape and material properties). While perceiving, the sense of touch may possibly be fused with other sensing modalities such as vision and auditory perception. Control commands are ultimately to be exerted in order to obtain the desired actions by the controller.

The last few decades have witnessed a tremendous progress in the development of tactile sensors, including diverse materials and methods explored to develop them, as extensively reported in \cite{dahiya2013directions,dahiya2010tactile,schmitz2011methods,kaltenbrunner2013ultra,yogeswaran2015new,khan2015technologies}. Research into the interpretation of tactile data to extract the information it conveys, particularly in the task of object recognition and localization, is nowadays also beginning to attract increasing attention. 
A comprehensive discussion on tactile sensory data interpretation is needed since many commonly used techniques, for example those adapted from the field of computer vision, may not always be suitable for tactile data.
This is because of the fundamentally different operating mechanisms of these two important sensory modalities.
%
%

To assist the advancement of the research in information extraction from tactile sensory data, this survey reviews the state-of-the-art in tactile perception of object properties, such as material identification, object recognition and pose estimation. In addition, work on perception that combines touch sensing with other sensing modalities, e.g., movement sensors and vision, are also studied.

The remainder of this paper is organized as follows: Available tactile sensors are first briefly introduced and compared in Section~\ref{sec:tactileprecption}. Research on material recognition via touch is then presented in Section~\ref{materialrecognition}. Works on tactile shape recognition and object pose estimation are reviewed on both local and global scales in Section~\ref{shaperecognition} and~\ref{localization} respectively. How vision and touch have been combined for object perception is discussed in Section~\ref{sensingintegration}. The last section concludes the paper and points to future directions in interpreting tactile data.

\section{Tactile perception as representation and interpretation of haptic sensing signals}
\label{sec:tactileprecption}
\subsection{Tactile sensing modalities}
\label{sec:tactsens}
\begin{figure}
	\centering	\includegraphics[width=.4\textwidth]{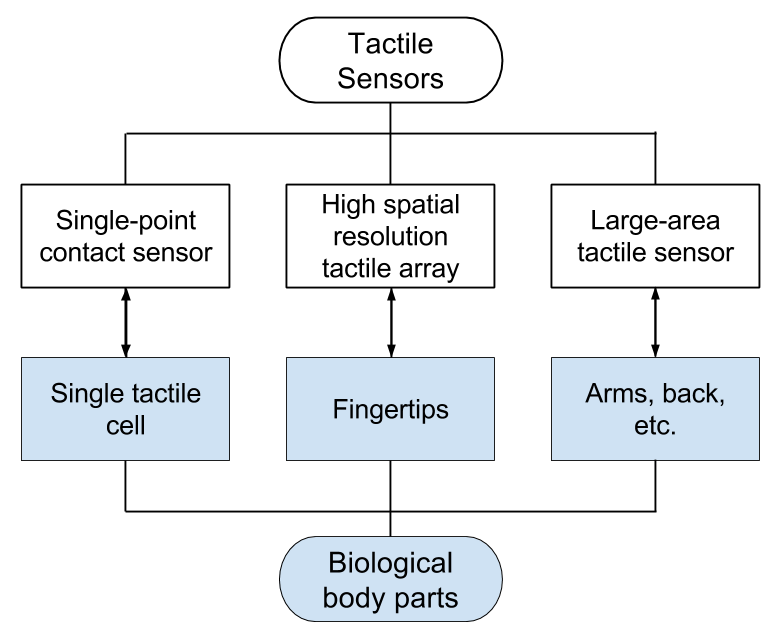}
	\caption{Tactile sensors of different types with the corresponding biological body parts in anatomy.}
	\label{fig:biovsrobot}
\end{figure}

As introduced in Section \ref{sec:intro}, tactile perception is the process of interpreting and representing touch sensing information to observe object properties.
In the hierarchy presented in Fig. \ref{fig:tactilesensingsystem}, it is placed a level above sensing, and provides useful, task-oriented information for planning and control \cite{siciliano2016springer}.
How tactile sensing information is interpreted and represented is closely linked with the type of hardware used and with the task to be fulfilled by the robot. Tactile sensing has been extensively reviewed in \cite{dahiya2013directions,dahiya2010tactile,argall2010survey,kappassov2015tactile} and the tactile sensors can be categorized in multiple manners, such as according to their sensing principles \cite{dahiya2011towards}, fabrication methods \cite{dahiya2013directions} and transduction principles \cite{dahiya2010tactile}. 
In this paper, 
we follow existing literature \cite{koiva2013highly,buscher2015flexible} and choose to categorize tactile sensors according to the body parts they are analogous to. As shown in Fig.~\ref{fig:biovsrobot}, tactile sensors in literature can be categorized into three types with respect to their spatial resolution, and an analogy can be made to corresponding parts of the human body. We provide a short description of each category in order to frame this review as follows.

\paragraph*{Single-point contact sensors, analogous to single tactile cells} This kind of sensor is used to confirm the object-sensor contact and detect force or vibrations at the contact point. Depending on the sensing modalities, single-point contact sensors can be categorized into: 1) force sensors for measuring contact forces, where a typical example is the ATI Nano 17 force-torque sensor; 2) biomimetic whiskers, also known as dynamic tactile sensors, for measuring vibrations during contact \cite{kroemer2011learning,fox2012tactile,kuchenbecker2006improving,mitchinson2014biomimetic,huet2017tactile}.
\paragraph*{High spatial resolution tactile arrays, analogous to human fingertips} Most research in tactile sensing is carried out using this type of tactile sensors \cite{kappassov2015tactile,dahiya2013directions}, and example prototypes are tactile arrays of 3$\times$4 tactile sensing elements based on fiber optics \cite{xie2013fiber}, tactile array sensors based on MEMS barometers \cite{tenzer2014feel} and fingertip sensors based on embedded cameras \cite{chorley2009development,sato2010finger,johnson2011microgeometry,yamaguchi2016combining}. There are also a multitude of commercial sensors available, such as RoboTouch and DigiTacts from Pressure Profile Systems (PPS)\footnote{\textcolor{blue}{www.pressureprofile.com/digitacts-sensors}}, tactile sensors from Weiss Robotics\footnote{\textcolor{blue}{www.weiss-robotics.com/en/produkte/tactile-sensing/wts-en/}}, Tekscan tactile system\footnote{\textcolor{blue}{www.tekscan.com/}}, BioTac multimodal tactile sensors from SynTouch\footnote{\textcolor{blue}{www.syntouchinc.com/}}. Among them, the most common are planar array sensors.
\paragraph*{Large-area tactile sensors, analogous to skin of human arms, back and other body parts} Unlike in fingertip tactile sensors, high spatial resolution is not essential in this type of tactile sensing. More importantly, they should be flexible enough to be attached to curved body parts of robots. The attention to developing this type of sensors has emerged in recent decades \cite{kaltenbrunner2013ultra,polat2015synthesis,hoffmann2017robotic}. Some researchers have developed this kind of tactile sensors for hands \cite{muscari2013real}, arms and legs \cite{mittendorfer2011humanoid,bartolozzi2016robots}, front and back parts \cite{kaboli2015humanoids} of humanoid robots. For a comprehensive review of large area and flexible tactile skins, the reader is referred to \cite{dahiya2013directions,khan2015technologies,dang2017printable,heidari2017bending}. 

\subsection{Tactile Perception}
Representations of tactile data are commonly either inspired by machine vision feature descriptors, where each tactile element is treated as an image pixel \cite{schneider2009object,pezzementi2011tactile}, biologically inspired\cite{Cannata2010}, or resort to dimensionality reduction\cite{heidemann2004dynamic,liu2012computationally,goger2009tactile}. Tactile information can be interpreted according to the desired function of the robot. Relevant information that can be extracted from sensing data include shape, material properties, and object pose. In this review, we focus on existing methods to extract these parameters, which are fundamental in the field of robot grasping and manipulation \cite{yousef2011tactile}, and have been used for the purpose of grasp control \cite{tegin2009demonstration}, slippage detection and prevention \cite{song2014efficient}, grasp stability assessment \cite{bekiroglu2011assessing}, among others. Other applications where tactile perception has been successfully applied range from haptic cues for Minimally Invasive Surgery (MIS) \cite{li2014multi}, creating interfaces for interactive games \cite{benali2004tactile} or medical training simulators \cite{coles2011role}, and assisting underwater robotic operations \cite{aggarwal2014object}.

Compared to the rapid development of tactile sensors, the interpretation of tactile sensors readings has not yet been fully taken into consideration. This is reflected by the small number of survey articles reported in the literature that focus on reviewing the computational intelligence methods applied in tactile sensing. The development of tactile sensors of increasing spatial resolution and fast temporal response provides an opportunity to apply state-of-the-art techniques of machine intelligence from multiple fields, such as machine learning, signal processing, computer vision and sensor fusion, in the field of tactile sensing. Accordingly, this paper explores recent advances in object shape and material recognition and pose estimation via tactile sensing and gives guidance towards possible future directions. In addition, it is also investigated how vision and touch sensing modalities can be combined for object recognition and pose estimation.

\section{Material recognition by tactile sensing}\label{materialrecognition}
The material properties of an object's surface are one of the most important cues that a robot requires for the sake of effectively interacting with its surroundings. Vision has been a popular approach to recognize object material \cite{liu2010exploring,sharan2013recognizing,sun2016recognising}. However, vision alone can only recognize a previously known surface material, and cannot, on its own, estimate its physical parameters.
In this respect, it is essential to utilize the sense of touch to identify the material properties. Surface texture (friction coefficients and roughness) and compliance are amongst the most crucial parameters for manipulating objects. Humans are extremely skilled in recognizing object material properties based on these cues \cite{lederman1990haptic} and a comprehensive review on human perception of material properties can be found in \cite{tiest2010tactual}. In robotics, researchers have endeavoured to enable a robot to identify the material properties at a level comparable to humans. 
These properties can be categorized into two different methods, i.e., surface texture based and object stiffness based.
\subsection{Surface texture based tactile material recognition}
Surface texture information can be extracted through the investigation of friction coefficients, roughness and micro-structure patterns of objects. The former two can be obtained using a force or tactile sensor sliding on the object surface, whereas the latter can be attained through tactile  images. The friction that arises at the contact while the sensor is sliding on an object's surface can be used to recognize the surface materials. Among the methods of this kind, the use of acoustic signals resulted from the friction to recognize the object materials is low-cost and requires limited computational power \cite{shan2017knock}. In \cite{roy1996surface}, a microphone is mounted on a robot leg which taps on the ground as the mobile robot moves, similar to the manner a blind person might tap his cane. The acoustic signature from tapping is then used to classify different floor materials. In \cite{edwards2008extracting}, an artificial finger equipped with a microphone (i.e., dynamic tactile sensor) is used to collect frictional sound data that are mapped to frequency domain through Fast Fourier Transform (FFT) to detect different textures. To recognize different texture surfaces, in \cite{kroemer2011learning} mean Maximum Covariance Analysis ($\mu$MCA) and weakly paired MCA (WMCA) are used to analyze the acoustic data after Fourier transform, which were collected by a dynamic tactile sensor. In \cite{johnsson2011sense}, a microphone-based texture sensor is employed and the textures can be classified using Self-Organizing Maps (SOMs). The use of acoustic signals has merits of low-cost and limited computational expense, however, ambient and motor noise may deteriorate the recognition performance. 

Strain gauges and force sensors are also used to detect vibrations during object-sensor interaction, in order to discriminate materials. By transferring raw data into the frequency domain by FFT, in \cite{jamali2010material}, different materials are classified based on surface textures by analyzing induced vibration intensities. In these works, an artificial finger is used with strain gauges and polyvinylidene fluoride films (PVDFs) embedded in silicone. In \cite{liu2012surface}, a dynamic friction model is applied to determine the surface physical properties while a robotic finger slides along the object surface with varying sliding velocity. The yarn tension data from a fabric sensor are also collected while the fingertip slides over objects in \cite{ho2012experimental}, with multiple computational intelligence methods for recognition being compared. In \cite{dallaire2014autonomous}, a tactile probe is proposed to measure the vibration signals while sliding to classify disks with different textures. This tactile probe is also used in \cite{giguere2011simple} to identify textures of different terrains. In \cite{sinapov2011vibrotactile}, artificial fingernails with three-axis accelerometers are used to scratch on surfaces and a frequency-domain analysis of the vibrations is done by using machine learning algorithms, i.e., \textit{k}NN and SVM. Similarly, vibrations are also used in \cite{romano2014methods}, in line with contact forces, to classify different surfaces. In addition to vibrations and force data, the output of proximity sensor is used to distinguish different surface textures in \cite{kaboli2014humanoids}.

Roughness is another important cue to discriminate between different object materials. In \cite{tanaka2003haptic}, object roughness is computed based on the variances of strain gauge signals using wavelet analysis. A microelectromechanical systems (MEMS) based tactile sensor is used to discriminate the roughness of object surfaces in \cite{oddo2011roughness}. Using a BioTac sensor in \cite{fishel2012bayesian}, Bayesian exploration is proposed to discriminate textures, which selects the optimal movements adaptively based on previous experience; good recognition performance is achieved for a large dataset of 117 textures.

The micro-structure patterns of objects can also be utilized to recognize object materials, usually with tactile array sensors. By using the camera-based GelSight sensor \cite{johnson2011microgeometry}, height maps of the pressed surfaces are treated as images to classify different surface textures using visual texture analysis \cite{li2013sensing}. Similarly, another camera-based tactile sensor TacTip is also used to analyze the object textures in \cite{winstone2013tactip}. In \cite{kim2005texture} a MEMS based tactile array sensor is employed to distinguish simple textures by using Maximum Likelihood (ML) estimation. The probability density function (PDF) of each texture type is created based on the mean and variance of the obtained tactile arrays and the textures are estimated by maximizing the PDFs. In \cite{shill2015tactile}, tactile images are also utilized to classify terrains, i.e., wood, carpet, clay and grass.

\subsection{Object stiffness based tactile material recognition} 
Object stiffness is also one of the critical material properties \cite{nanayakkara2016stable}. By using a BioTac sensor, the object compliance (the reciprocal of stiffness) can be estimated either using the contact angle of the fingertip \cite{xu2013tactile} or investigating BioTac electrode data \cite{su2012use,hoelscher2015evaluation,chu2013using}. In \cite{windau2010inertia} a robot leg equipped with an accelerometer is employed to actively knock on object surfaces and by analysing the sensor data, the hardness, elasticity and stiffness of the object can be revealed. In recent works \cite{yuan2016estimating,yuan2017shape}, the hardness of objects can also be estimated by processing the tactile image sequences from a GelSight sensor. In \cite{decherchi2011tactile}, multiple computational algorithms are applied to classify various materials based on mechanical impedances using tactile data and it is found that SVM performs best. In \cite{drimus2014design}, by using image moments of tactile readings as features, dynamic time warping is used to compare the similarity between time series of signals to classify objects into rigid and deformable. A force sensor is used to test the mechanical impedances of materials in terms of shear modulus, locking stretch and density in \cite{sangpradit2011finite}. A multi-indenting sensing device is proposed in \cite{faragasso2015multi} to measure the stiffness of the examined object, i.e., phantom soft tissue used in this work.


In summary, object materials can be recognized by using different touch sensing cues based on surface textures, vibrations and mechanical impedances, summarized in Table~\ref{tab:materialrecognitionsummary} with a comparison of their pros and cons.

\begin{table*}[tbh!]
	\caption{A summary of material recognition methods with touch sensing}
	\centering
	\label{tab:materialrecognitionsummary}
	\begin{tabular}{m{1.5cm}| m{1.5cm}|m{2cm}| m{3cm}| m{3cm}| m{3cm}| m{0.5cm}}
		\hline
		Methods & Motions & Data types & Advantages & Disadvantages & Applicable sensors & Ref. \\  \hhline{=|=|=|=|=|=|=}
		\multirow{5}{*}{Texture based} & Sliding, tapping, scratching & Frictions, acoustic data, acceleration, vibrations & Low cost and limited computational expenses & Interaction actions (sliding, tapping, scratching) may damage objects & Dynamic tactile sensors, force sensors, strain gauges and PVDF sensors & \cite{johnsson2011sense,jamali2011majority,sinapov2011vibrotactile} \\
		\hhline{~------}
		& Imprint & Tactile images  & Micro-structure patterns of object textures can be captured in one tactile image & Low resolution of tactile sensors may cause difficulties for processing & Planar array sensors; potentially applied to curved tactile sensors & \cite{li2013sensing,kim2005texture}\\
		\hline
		Mechanical impedances based & Squeezing, knocking, pressing  & Force variances, tactile images, acceleration  & Limited computational costs; complementary to other information  & Interaction actions (knocking, squeezing) may damage objects & Accelerometer-based sensors, force sensors and tactile array sensors & \cite{windau2010inertia,drimus2011classification,kaboli2014humanoids}  \\
		\hline
	\end{tabular}
\end{table*}

\section{Tactile object shape perception} \label{shaperecognition}
Object shape perception is the ability to identify or reconstruct the shape of objects. The goals of shape perception vary in different tasks, from capturing the exact shape, like getting the point cloud of the object, to classification of shape elements or overall profiles. This capability is crucial for robots to perform multiple tasks such as grasping and in-hand manipulation. The more complete information is obtained about the object's shape, the more capable the robot will be to plan and execute  grasping trajectories and manipulation strategies. Research into shape recognition has been dominated by vision based methods \cite{lowe1999object,felzenszwalb2005pictorial}. However, visual shape features cannot be observed when vision is occluded by hand or in poor illumination conditions. In contrast, tactile object shape perception is not affected by such factors and can observe detailed shapes by sensor-object interactions. The surge of high-performance tactile sensors gives rise to the emergence and rapid spread of algorithms to recognize object shapes via touch.

\begin{figure}
	\centering
	\begin{subfigure}[b]{0.24\textwidth}
		\centering
		\includegraphics[height=1.61cm,width=1.61cm]{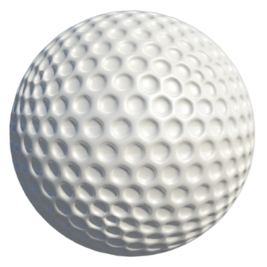}
		\caption{}
		\label{fig:barshape}
	\end{subfigure}%
	\hfill
	\begin{subfigure}[b]{0.24\textwidth}
		\centering
		\includegraphics[height=1.61cm,width=1.61cm]{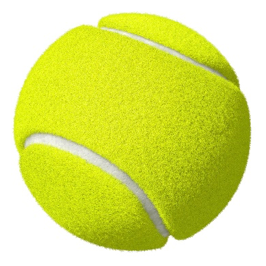}
		\caption{}
		\label{fig:elipseshape}
	\end{subfigure}%
	\hfill
	\caption{Two ball shapes with different local shapes: one is a golf ball and the other is a tennis ball. At a global scale, both of them are balls. At a local scale, (a) has small pits whereas (b) has curvilinear shapes. } 
	\label{fig:ballshape}
\end{figure}

The perception of object shapes can be done on two scales, i.e., local and global, as illustrated in Fig.~\ref{fig:ballshape}. The former can be revealed by a single touch through tactile image analysis. It is analogous to the human cutaneous sense of touch, which is localized in the skin. The latter reflects the contribution of both cutaneous and kinaesthetic feedback, e.g., contours that extend beyond the fingertip scale. In this case, intrinsic sensors, i.e., proprioceptors in joints, are also utilized to acquire the position and movement of the fingers/end-effectors that are integrated with local features to recognize the objects. Here the kinaesthetic cues are similar to human proprioception that refers to the awareness of the positions of the body parts.

\subsection{Local shape recognition}
\label{localshapeclassification}
In terms of identifying local shapes, recent years saw a trend to treat pressure patterns as images, thereby extracting features based on the pressure distribution within the images \cite{pezzementi2011tactile}. The increasing spatial resolution and spatio-temporal response enable tactile sensors to demonstrate the ability to serve as an ``imaging" device. A large number of researchers have applied feature descriptors from vision such as image moments \cite{pezzementi2011tactile,corradi2015bayesian} to tactile data to represent local shapes. However, there are differences between vision and tactile imaging as listed in Table~\ref{tab:visionvstactile}. For vision, the field of view (FoV) is large and global since multiple objects can exist in a single camera image. A great amount of features can be obtained from one single image and it is relatively easy to collect data by cameras. On the other hand, it requires high computational resources to process the visual data; there are several fluctuation factors that affect the performance of extracted features, e.g., scaling, rotation, translation and illumination. Here scaling is caused by the distance from cameras to observed objects. In contrast, for tactile sensing, the FoV is small and local as direct sensor-object interactions need to be made. And the information available in one reading is limited due to the low sensor resolution (for instance, Weiss tactile sensor of 14$\times$6 sensing elements compared to a typical webcam of 1280$\times$1024 pixels), especially in terms of revealing the appearance and geometry of objects. Compared to vision, it is relatively expensive to collect tactile data, in terms of sensors and robot components but there is less effort involved, as it requires less computations to process the tactile data. In addition to these properties, features can be extracted from the tactile data that include surface texture \cite{jamali2011majority}, mechanical impedance \cite{xu2013tactile} and local detailed shapes. In tactile imaging, the influence of scaling is removed as the real dimension and shape of the interacted object can be mapped to the tactile sensor directly, whereas the impact of rotation, translation and ``illumination" remains. Here, ``illumination" refers to different impressions of object shapes caused by forces of various magnitudes and directions, similar to the variety of light conditions in vision.

\begin{table}
	\centering
	\begin{threeparttable}
		\centering
		\caption{Comparison of vision and tactile sensing}
		\label{tab:visionvstactile}
		\begin{tabular}{l|m{0.7cm}|m{0.6cm}|m{0.7cm}|m{0.7cm}|m{2.7cm} }
			\hline
			Modality & FoV & Info.  & Compl. & Compu. & Invariance \\
			\hhline{=|=|=|=|=|=}
			Visual & Global & Rich & Low & High & Scaling, rotation, translation, illumination \\
			\hline 
			Tactile & Local & Sparse & High & Low & Rotation, translation, ``illumination" \\
			\hline 
		\end{tabular}
		\begin{tablenotes}
			\footnotesize
			\item Note: FoV: Field of View; Info.: Information; Compl.: Complexity to collect data; Compu.: Computation. ``illumination": different impressions caused by forces of various magnitudes and directions.
		\end{tablenotes}
	\end{threeparttable}
\end{table}

\subsubsection{Shape descriptors for tactile object recognition}

As raw tactile readings are to some extent redundant, features are utilized to represent the collected tactile data. The paradigm is to extract features based on the pressure distribution in tactile arrays and then feed the extracted features into classifiers to build a training model that can be used to recognize test object shapes. According to the descriptors used, methods for local shape recognition can be categorized as follows:

\paragraph*{Raw tactile readings as features} This method avoids the feature extraction process and is easy to implement \cite{jimenez1997featureless,liu2012tactile,dang2011blind}. However, on the other hand it is sensitive to pattern variations in positions, orientations and sizes. In \cite{schneider2009object}, the columns of each tactile matrix are concatenated to form a vector that is directly treated as a descriptor, i.e., a ``do-nothing" descriptor. The feature is sensitive to the pose variances of objects, as a result, a single object is assigned with multiple identities if it is placed in different orientations to the robotic gripper. In \cite{pezzementi2011tactile}, the method using a ``do-nothing" descriptor is taken as a baseline and shows worse performance compared to the other methods. In \cite{martinez2013active}, the tactile readings of iCub fingertip with 12 taxels are taken to classify regions that the fingertip taps into edges, plane and air.

\paragraph*{Statistical features} It is effortless to obtain statistical features but the extracted features cannot be guaranteed to be useful. In \cite{schopfer2009using}, statistical features are utilized, e.g., maximum, minimum and mean pressure values of each tactile reading, and positions of the center of gravity. As a result, the statistical features of 52 different types form a 155 dimensional feature vector. As the information that resides in some features is redundant, an entropy based method is applied to investigate the usefulness of the features but as a result low recognition rate of only around 60\% is achieved. The statistical features have also been applied in other applications. In \cite{tawil2012interpretation,tawil2011touch}, touch attributes, e.g., pressure intensity and area of contact, are taken as features to classify the touch modalities. In \cite{chitta2011tactile}, the internal states of bottles and cans are estimated using the designed statical features of the tactile data.

\paragraph*{Descriptors adapted from computer vision} In these methods, tactile arrays are treated as images and thus vision descriptors can be applied and adapted. Several researchers took image moments as feature descriptors \cite{pezzementi2011tactile,corradi2015bayesian,russell2000object,bekiroglu2011assessing,drimus2014design}. For a tactile reading $f(x,y)$, the image moment $ m_{pq} $ of order $p+q$ can be calculated as follows:
\begin{equation}
m_{pq}=\sum_{x}\sum_{y}x^{p}y^{q}f(x,y),
\end{equation}
where $p$ and $q$ stand for the order of the moment, $x$ and $y$ stand for the horizontal and vertical positions of the cell in the tactile image, respectively. In most cases, the moments of order up to 2, i.e., $ (p+q)\in\{0,1,2\} $, are computed and other properties are based on them, like Hu's moments used in \cite{pezzementi2011tactile} and Zernike moments used in \cite{corradi2015bayesian}.  For example, in \cite{schmid2008opening}, a tactile reading is approximated by an ellipse whose principal axes are represented by the second order central moments. Some other vision descriptors are also applied in tactile sensing like regional descriptors used in \cite{khasnobish2014object}. The Scale Invariant Feature Transform (SIFT) descriptor is created based on image gradients \cite{lowe2004distinctive} that has been proved robust to cope with object pose variations in vision applications. This is also particularly useful for tactile object recognition, since the touches could introduce unexpected object rotation and translation. SIFT has been explored in \cite{pezzementi2011tactile,luo2014rotation,luonovel,luo2015tactile} for tactile recognition and a good performance can be achieved. As both visual images and tactile readings are present in numerical matrices, many other vision descriptors \cite{mikolajczyk2005performance,gauglitz2011evaluation} also have the potential to be applied in tactile sensing, e.g., Shape Context \cite{belongie2002shape}, SURF \cite{bay2006surf} and  3D descriptor SHOT \cite{tombari2010unique}.

Vision descriptors have also been applied in other tactile applications \cite{nagatani2012can}. Inspired by the similarities between tactile patterns and grey-scale images, in \cite{ji2011histogram} tactile data are transformed into histograms as structured features to discriminate human-robot touch patterns. In \cite{schurmann2011modular}, the pose of objects placed on the sensor, i.e., the orientation of a cup handle in this work, is  estimated by applying Hough transform to the impressed tactile profiles. In \cite{wong2014spatial}, the edge orientation is estimated with a BioTac sensor by using a support vector regression (SVR) model. Image moments have also been widely used in these applications. In \cite{bekiroglu2011assessing,bekiroglu2011learning}, image moments are utilized to represent the acquired tactile data, applied in grasp stability analysis. In a more recent work \cite{bekiroglu2013probabilistic}, in addition to the image moments mentioned above, more haptic features including the 3D version of image moments and average normal vector are taken into consideration for the same task, together with features of other sensory streams, i.e., vision and proprioception. In \cite{licontrol}, both Hough transform and image moments are employed and compared to predict the orientation of an object edge so as to perform tactile servoing.

The GelSight tactile sensor has an extra-high resolution of 320$ \times $240 and provides the ability of using high level vision descriptors  \cite{li2014localization}. It consists of a camera at the bottom and a piece of clear elastomer on the top.  The elastomer is coated with a reflective membrane, which deforms to take the surface geometry of the objects on it. The deformation is then recorded by the camera under illumination from LEDs that project from various directions. With the use of GelSight sensor, a multi-scale Local Binary Pattern (LBP) descriptor is developed and used to extract both micro- and macro-structures of the objects from obtained tactile readings/images for texture classification \cite{li2013sensing}.

\paragraph*{PCA-based features} Principal Component Analysis (PCA) can be applied to tactile readings and the acquired principal components (PCs) are taken as features. It can reduce the redundancy of tactile data and is easy to implement but it lacks physical meaning \cite{gorges2010haptic}. In \cite{heidemann2004dynamic}, a $ 16 \times 16 $ tactile reading is projected onto a feature space of a lower dimension and an iterative procedure is taken to compute PCs with the largest eigenvalues.  In \cite{liu2012computationally}, the pressure distribution is defined as $ M=[x \ y \ p]^T $, where $x$, $y$ are location coordinates in the sensor plane and $p$ is the pressure in this location. By covariance analysis of $M$, the resultant eigenvector lengths, principal axis direction and shape convexity are taken as features to recognize local shapes. The kernel PCA-based feature fusion is used in \cite{liu2014low} to fuse geometric features and Fourier descriptors (based on Fourier coefficients) to better discriminate objects. In \cite{li2014learning}, PCA is applied to reduce the dimensionality of tactile readings. The obtained tactile features are then used for grasp stability assessment and grasp adaptation. In \cite{scho2007acquisition}, PCA is also employed to extract features from tactile readings that is applied for object pose estimation\cite{bimbo2016hand}.
\paragraph*{Self-organizing features} The aforementioned features are predefined and hand-crafted and they are fed into shallow classifiers, such as single-layer neural networks, \textit{k}NN, and SVM. Methods of this type can be easily implemented but they can restrict the representation capability to serve different applications and may only capture insignificant characteristics for a task when using hand-crafted features \cite{madry2014st}. In contrast, there is no need to define the feature representation format a-priori when using multilayer/deep architecture methods to learn self-organizing features from raw sensor data. Soh et al. \cite{soh2012online} developed an on-line generative model that integrates a recursive kernel into a sparse Gaussian Process. The algorithm iteratively learns from temporal tactile data and produces a probability distribution over object classes, without constructing hand-crafted features. They also contribute discriminative and generative tactile learners \cite{soh2014incrementally} based on incremental and unsupervised learning. In \cite{madry2014st} and \cite{molchanov2016contact}, unsupervised hierarchical feature learning using sparse coding is applied to extract features from sequences of raw tactile readings, for grasping and object recognition tasks. In \cite{schmitz2014tactile} denoising autoencoders with dropout are applied in tactile object recognition and a dramatic performance improvement of around 20\% is observed in classifying 20 objects compared to using shallow neural networks and supervised learning. In \cite{cao2016efficient}, the randomized tiling Convolutional Network (RTCN) is applied for the feature representation in tactile recognition and can achieve an extremely good recognition rate of 100\% for most of the datasets tested in the paper. In \cite{liu2016object}, a joint kernel sparse coding model is  proposed for the classification of tactile sequences acquired from multiple fingers. As new techniques in deep learning and unsupervised learning emerge and grow rapidly in recent years \cite{bengio2013representation}, it is promising to apply more such algorithms to acquire self-organizing features in tactile sensing. On the other hand, though deep learning shows tremendous promise for the object recognition tasks, online application of such a computationally intensive process is difficult and it is hard to tune the parameters of deep architectures; also, the complexity of these deep architectures translate into difficulties in the introspection and physical interpretation of the resulting model.

\subsubsection{Discussions of local shape descriptors}
A summary of the discussed tactile features is given in Table~\ref{tab:tactilefeaturesummary}, with discussions and comparison of pros and cons, and guidance on the selection of tactile sensors. In the state-of-the-art literature, vision based descriptors that take tactile readings as images are widely employed and will also be the mainstream in extracting features for tactile object recognition and other applications. There is another trend to employ unsupervised learning and deep architectures to learn self-organizing features from raw tactile readings as an increasing number of such algorithms are being developed \cite{bengio2013representation}. In addition to shape recognition, the various descriptors discussed here have also been used to identify different contact patterns in several other tasks that can be grouped as grasp stability assessment \cite{madry2014st,bekiroglu2013probabilistic,bekiroglu2011assessing,dang2011blind,bekiroglu2011learning,hyttinen2015learning}, identification of touch modalities \cite{kaboli2015humanoids,ji2011histogram,tawil2012interpretation}, object pose estimation \cite{schurmann2011modular}, slip detection \cite{nagatani2012can}, learning system states during in-hand manipulation \cite{stork2015learning}, contour following \cite{martinez2013active}, tactile servoing \cite{licontrol}, surface texture recognition \cite{li2013sensing}, localization and manipulation in assembly tasks \cite{li2014localization}, and grasp adaptation \cite{li2014learning}. 
They can also be applied in other applications in the future research.

\begin{table*}[htbp]
	\caption{A summary of previously studied tactile features}
	\centering
	\label{tab:tactilefeaturesummary}
	\begin{tabular}{m{1.5cm}| m{5cm}| m{5cm}| m{2.5cm}| m{1.5cm}}
		\hline
		Feature type  & Advantages & Disadvantages & Applicable sensors & Ref. \\  \hhline{=|=|=|=|=}
		Raw tactile readings & A ``do-nothing" descriptor; easy to implement; applied to any type of tactile data & Lack of physical meaning; redundant; sensitive to variations of contact forces and poses & Planar array sensors; curved tactile sensors & \cite{schneider2009object,dang2011blind,martinez2013active} \\
		\hline
		Statistical features & Easy to be computed; based on statistics; can be applied to any type of tactile data & Lack of physical meaning; redundant; sensitive to force and pose variance; hand-crafted & Planar array sensors; curved tactile sensors & \cite{schopfer2009using,tawil2012interpretation,tawil2011touch,kaboli2014humanoids} \\
		\hline
		Descriptors adapted from vision & Extract distinct features; invariant to force and pose variance; can be used to share information with vision & Some of the features are of high dimension compared to original reading; hard to design; predefined and hand-crafted & Planar array sensors; low curvature sensors & \cite{pezzementi2011tactile,corradi2015bayesian,russell2000object,drimus2014design,bekiroglu2013probabilistic,bekiroglu2011learning,schurmann2011modular} \\
		\hline
		PCA-based features & Low dimensionality; statistics based; easy to implement & Lack of physical meaning; sensitive to force and pose variance & Planar array sensors; curved tactile sensors & \cite{li2014learning,liu2012computationally,bimbo2016hand} \\
		\hline
		Self-organizing & No need to define the feature representation format a-priori; learn features from raw data & Lack of physical meaning; high computational complexity; hard to tune parameters & Planar array sensors; curved tactile sensors & \cite{soh2012online,madry2014st,schmitz2014tactile,soh2014incrementally,cao2016efficient} \\
		\hline
	\end{tabular}
\end{table*}

\subsection{Global shape perception}
\label{globalshaperecognition}
The methods to recognize or reconstruct the global shape of objects with tactile sensing can be grouped into three categories with respect to the sensing inputs: 1) methods using the distributions of contact points obtained from single-point contact force sensors or tactile sensors; 2) methods based on analysing the pressure distributions in tactile arrays; 3) methods of combining both tactile patterns and contact locations. Here the global shape refers to the overall shape of objects, especially contours that extend beyond the fingertip scale.

\paragraph*{Points based recognition} The methods of this type often employ techniques from computer graphics to fit the obtained cloud of contact points to a geometric model and outline the object contour. This method was widely used by early researchers due to the low resolution of tactile sensors and prevalence of single-point contact force sensors \cite{grimson1984model,allen1989haptic,charlebois1999shape,okamura2001feature}. In \cite{allen1989haptic} resultant points from tactile readings are fit to super-quadric surfaces to reconstruct unknown shapes. In a similar manner, relying on the locations of contact points and hand pose configurations, a polyhedral model is derived to recover object shapes in \cite{casselli1995robustness}. These approaches are limited as objects are usually required to be fixed and stationary. Different from the point cloud based approaches, a non-linear model-based inversion is proposed in \cite{fearing1991using} to recover surface curvatures by using a cylindrical tactile sensor. In more recent works \cite{ibrayev2005semidifferential,jia2006surface,jia2010surface}, the curvatures at curve intersection points are analyzed and thus a patch is described through polynomial fitting; in \cite{abraham2017ergodic}, estimation of nonparametric shapes is demonstrated using binary sensing (collision and no collision) and ergodic exploration.

In some other works tactile sensors are utilized to classify objects by taking advantage of the spatial distribution of the object in space. In \cite{pezzementi2011object} an object representation is constructed based on mosaics of tactile measurements, in which the objects are a set of raised letter shapes. In this work, the object recognition is regarded as a problem of estimating a consistent location within a set of object maps and thus histogram and particle filtering are used to estimate possible states (locations). 
A descriptor based on the histogram of triangles generated from three contact points was used in \cite{zhang2016triangle} to classify 10 classes of objects. This descriptor is invariant to object movements between touches but requires a large number of samples (grasps) to accurately classify the touched object.
Kalman filters are applied in \cite{meier2011probabilistic} to generate 3D representations of objects from contact point clouds collected by tactile sensors and the objects are then classified by the Iterative Closest Point (ICP) algorithm. A similar method is employed in \cite{aggarwal2015haptic}, for haptic object recognition in underwater environments. Through utilizing these methods, arbitrary contact shapes can be retrieved, however, it can be time consuming when investigating a large object surface as excessive contacts are required for recognizing the global object shape.

\paragraph*{Tactile patterns based recognition} Another approach is to recognize the contact shapes using pressure distribution within tactile arrays. As a result of the increasing performance of tactile sensors, this approach has become increasingly popular in recent years. Various methods to recognize the local contact shapes have been reviewed and discussed in Section~\ref{localshapeclassification}. In terms of recognising the global object shape by analysing pressure distributions in tactile images collected at different contact locations, however, a limited number of approaches are available. One popular method is to generate a codebook of tactile features and use it to classify objects and a particular paradigm is the Bag-of-Features (BoF) model \cite{madry2014st,schneider2009object,pezzementi2011tactile}. The BoF model originates from the Bag-of-Words (BoW) model in natural language processing for text classification and has been widely utilized in the field of computer vision \cite{nowak2006sampling}, thanks to the simplicity and power of the model. Inspired by the similar essence of vision and tactile sensing, Schneider et. al. \cite{schneider2009object} first applied the BoF model in tactile object recognition. In this framework, local contact features extracted from tactile readings in the training phase are clustered to form a dictionary and cluster centroids are taken as ``codewords". Based on the dictionary, each tactile feature is assigned to its nearest codeword and a fixed length feature occurrence vector is generated to represent the object. It is easy to implement and can achieve an appropriate performance \cite{schneider2009object,pezzementi2011tactile}, however, only local contact patterns are taken and the distribution of the features in three-dimensional space is not incorporated.
\paragraph*{Object recognition based on both sensing modalities} For humans, the sense of touch consists of both kinaesthetic and cutaneous sensing and these two sensing modalities are correlated \cite{lederman2009haptic}. Therefore, the fusion of the spatial information and tactile features could be beneficial for the object recognition tasks. This combination has already been proved to improve recognition capabilities in teleoperation experiments with human subjects both in identifying curvatures \cite{pattichizzo2013towards} and estimating stiffness \cite{pacchierotti2013improving}.
In \cite{mcmath1991tactile,petriu1992active}, a series of local ``tactile probe images" is assembled and concatenated together to obtain a ``global tactile image" using 2D correlation techniques with the assistance of kinaesthetic data. However, in this work tactile features are not extracted whereas raw tactile readings are utilized instead, which would bring high computational cost when investigating large object surfaces. In \cite{johnsson2007neural}, three different models are proposed based on proprioceptive and tactile data, using Self-Organising Maps (SOMs) and Neural Networks. In \cite{gorges2010haptic}, the tactile and kinaesthetic data are integrated by decision fusion and description fusion methods. In the former, classification is done with two sensing modalities independently and recognition results are combined into one decision afterwards. In the latter, the descriptors of kinaesthetic data (finger configurations/positions) and tactile features for a single palpation are concatenated into one vector for classification. In other words, the information of the positions where specific tactile features are collected is lost. In both methods, the tactile and kinaesthetic information is not fundamentally linked. In a similar manner, in \cite{navarro2012haptic} the tactile and kinaesthetic modalities are fused in a decision fusion fashion. Both tactile features and joint configurations are clustered by SOMs and classified by ANNs separately and the classification results are merged to achieve a final decision. In a more recent work \cite{spiers2016single}, the actuator positions of robot fingers and force values of embedded TakkTile sensors form the feature space to classify object classes using random forests but there are no exploratory motions involved, with data acquired during a single and unplanned grasp. In \cite{luo2016iterative}, an algorithm named Iterative Closest Labeled Point (iCLAP) is proposed to recognize objects using both tactile and kinaesthetic information that has been shown to outperform those using either of the separate sensing modalities. In general, it is still an open question how to link both sensor locations and tactile images.


\begin{table*}[htbp]
	\caption{Methods for global shape recognition}
	\centering
	\label{tab:globalrecognition}
	\begin{tabular}{m{1.2cm}|m{4cm}|m{3.5cm}|m{3.5cm}|m{2.3cm}}
		\hline
		Modalities & Recognition methods & Advantages & Disadvantages & Applicable sensors \\  \hhline{=|=|=|=|=}
		Contact points & Graphical models (point clouds) \cite{casselli1995robustness}; polynomial fitting (surface curvatures) \cite{jia2010surface}; filtering (spatial distribution) \cite{pezzementi2011object,meier2011probabilistic} &  Arbitrary shapes can be retrieved; object graphical models can be built; spatial distribution is revealed & Time consuming when investigating large surfaces; excessive contacts required; local features are not revealed & Single-point contact sensors; planar/curved tactile sensors \\
		\hline
		Tactile patterns & Bag-of-Features \newline \cite{luonovel,madry2014st,schneider2009object,pezzementi2011tactile} & Easy to implement; local shape features are employed & The distribution of the features in 3D space is not incorporated &  Planar sensors or low curvature \\
		\hline
		Both sensing modalities & Image stitching \cite{mcmath1991tactile,petriu1992active}; Decision fusion \cite{gorges2010haptic,navarro2012haptic}; Description fusion \cite{gorges2010haptic} & Both kinaesthetic and cutaneous cues are included & Hard to associate local patterns and kinaesthetic data; bring additional computational cost & Planar sensors or low curvature \\
		\hline 
	\end{tabular}
\end{table*}

In summary, different strategies have been taken to recover the global shape of objects in the view of proprioception and tactile sensing, as summarized in Table~\ref{tab:globalrecognition} with a comparison of pros and cons of different methods. Some researchers take advantage of the distributions of contact points in space, e.g., points based methods, whereas some others utilize local patterns only, e.g., BoF framework; and it is also expected to achieve a better perception by combining both distributions of contact points in space and local contact patterns.

\section{Pose estimation via touch sensing}\label{localization}
Effective object manipulation requires accurate and timely estimation of the object pose.
This pose is represented by the object's position and orientation with respect to the robot end-effector or to a global coordinate frame.
Even small errors the estimate of the object's location can lead to incorrect placement of the robot fingers on the object, generate wrong assumptions on grasp stability and compromise the success of a manipulation task.
In fact, in-hand manipulation is, by definition, the task of changing an object's pose from an initial to a final configuration \cite{hertkorn2013planning}.
Thus, robust, accurate and fast perception of an object's pose must be a crucial part of any sophisticated grasping and manipulation system.

The most common means in robotics to estimate an object's pose is using computer vision.
However, when a robot approaches the object to be manipulated, it creates occlusions and vision cannot be relied upon.
To cope with this problem, tactile sensing has been used to assist a robot in determining the pose of a touched object, either on its own or in combination with a vision system.
In this review we classify existing techniques according to the sensing inputs: single-point contact sensor and tactile sensing arrays, on their own or used together with a vision system (contact-visual and tactile-visual), as shown in Table~\ref{tab:slam}.


\paragraph*{Single-point contact based} Due to the poor performance of tactile sensors, most early works tend to use single-point contact sensors, i.e., force sensors and dynamic tactile sensors, for localizing the objects or features. 
Early work on finding an object's pose used only angle and joint-torque sensing and used an interpretation tree that contained possible correspondences between object vertices to fingers\cite{siegel1991finding}.
In \cite{gadeyne2005bayesian}, a force-controlled robot is used to localize objects using Markov Localization that is applied in a task of inserting a cube (manipulated object) into a corner (environment object) by a manipulator. This compliant motion problem is also compared with the data association  \cite{montemerlo2003simultaneous}, i.e., assigning measurements to landmarks, and global localization with different models in Simultaneous Localization And Mapping (SLAM) for mobile robotics. In both cases, Bayesian based approaches can provide a systematic solution to estimate both models and states/parameters simultaneously. In \cite{schaeffer2003methods}, a set of algorithms are implemented for SLAM during haptic exploration but only simulation results are presented. 

Particle filtering is popular in (optical/acoustic-based) robot localization problems and may merit further investigation in tactile sensing, where objects could be modelled as clouds of particles distributed according to iteratively updated probability distributions. In \cite{corcoran2010measurement}, particle filtering is applied to estimate the object pose and track the hand-object configurations, especially in the cases where objects are possibly moving in the robot hand. Also using particle filtering, small objects, e.g., buttons and snaps, are localized and manipulated in flexible materials such as fabrics that are prone to move during robot manipulations in \cite{platt2011using}. 
Improvements of the particle filter algorithm have been presented to address the problem of localizing an object via touch.
In \cite{petrovskaya2006bayesian,petrovskaya2007touch,petrovskaya2011global}, this novel particle filter, named Scaling Series, has each particle representing not a single hypothesis, but a region in the search space which is sequentially refined. Besides, annealing is used to improve sampling from the measurement model.
This method was tested using a robot manipulator equipped with a 6D force/torque sensor. It is applied in two scenarios: 1) to localize (estimate positions and orientations), grasp and pick up a rectangular box; 2) to grasp a door handle. 
A method that includes memory from past measurements and a ``scaled unscented transformation'' is performed on the prediction step was presented in \cite{vezzani2016memory}.

A ``haptic map" is created from the proprioceptive and contact measurements during the training that is then used to localize the small objects embedded in flexible materials during robot manipulation. By using dynamic tactile sensors inspired by rat whiskers, the grid based SLAM is introduced in  \cite{fox2012tactile} to navigate a robot with touch sensors by deriving timing information from contacts and a given map about edges in a small arena, in which particle filters are also used. In \cite{yu2015shape}, SLAM is applied to recover the shape and pose of a movable object from a series of observed contact locations and contact normals with a pusher. Analogous to SLAM in mobile robotics, the object is taken as a rigid but moving environment and the pusher is taken as a sensor to get noisy observations of the locations of landmarks. However, compared to the exploration using visual feedback, tactile exploration is challenging in the sense that touch sensing is intrusive in nature, that is, the object (environment) is moved by the action of sensing.

\begin{table}[htbp]
	\centering
	\begin{threeparttable}
		\caption{Different sensing inputs for object pose estimation}
		\label{tab:slam}
		\begin{tabular}{m{1.7cm}| m{2.7cm}| m{1.5cm}| m{0.8cm}}
			\hline Types & Sensors involved & Information & Example \\  
			\hhline{=|=|=|=}
			Visual & Cameras, lasers & Global & \cite{davison2007monoslam} \\ 
			\hline Contact & Single-point contact sensors & Local & \cite{petrovskaya2011global} \\ 
			\hline Tactile & Tactile sensors & Local & \cite{li2014localization} \\ 
			\hline Contact-Visual & Cameras, single-point contact sensors & Global+local & \cite{Bhattacharjee2015combining} \\
			\hline Tactile-Visual & Cameras, tactile sensors & Global+local & \cite{luo2015localizing} \\ 
			\hline
		\end{tabular}
		\begin{tablenotes}
			\footnotesize
			\item Note: Laser scanners and cameras (widely used in mobile robots) collect global information about the environment, whereas the contact sensors (single-point contact sensors and tactile sensors) provide local information.
		\end{tablenotes}
	\end{threeparttable}
\end{table}

\paragraph*{Tactile-based} The increasing performance of tactile sensors provides the feasibility to localize objects or object features in robot hand using the information derived from tactile arrays. With the high-resolution GelSight sensor, collected tactile images can be localized \cite{li2014localization} within a height map via image registration to help localize objects in hand. A height map is first built based on the collected tactile readings. The keypoints are then localized from both the map and incoming tactile measurements. After that, feature descriptors are extracted from both and matched. In this manner, the pose of an assembly part in the robot hand can be estimated.
Similarly to the way that local geometric features can be extracted using PCA, these features were used to determine an object's pose.
In \cite{bimbo2015global} a Monte Carlo method was used to find an object pose where the local geometry of the object at the contact location matched the PCA features obtained from the tactile sensor.
Another approach relied on the fact that, when the robot is in contact with the object, the possible locations of the object must lie inside a contact manifold, a novel particle filter was developed that was both faster and more accurate than a standard particle filter \cite{koval2015pose}.
\paragraph*{Contact-Visual based} Vision and tactile sensing share information of how objects are present in 3D space in the form of data points. For vision, a mesh can be generated based on point cloud from 3D cameras or laser scanners that can also be obtained by contact sensors (in this case, mostly single-point contact sensors). In \cite{hebert2011fusion}, localizing object within hand is treated as a hybrid estimation problem, by fusing data from stereo vision, force-torque and joint sensors. Though reasonable object localization can be achieved, tactile information on the fingertips is not utilized, being however promising to obtain better performance of object localization by incorporating additional information. Based on the assumption that visually similar surfaces are likely to have similar haptic properties, vision is used to create dense haptic maps efficiently across visible surfaces with sparse haptic labels in \cite{Bhattacharjee2015combining}.
Vision can also provide an approximate initial estimate of the object pose that is then refined by tactile sensing using local\cite{honda1998realtime,bimbo2013combining} or global optimization\cite{bimbo2015global}.

\paragraph*{Tactile-Visual based} Vision and tactile sensing of humans have also been found to share the representations of objects \cite{lacey2007vision}. To put it in another way, there are some correspondences, e.g., prominent features, between vision and tactile sensing while observing objects. In \cite{luo2015localizing}, it is proposed to localize a tactile sensor in a visual object map by sharing similar sets of features between visual map and tactile readings. It is treated as a probabilistic estimation problem and solved in a framework of recursive Bayesian filtering, where tactile patterns are for local information and vision is to provide a global map to localize sensor contact on the object.


In summary, the localization of objects during manipulation using tactile sensing remains an open problem.
While the most popular approaches rely on techniques similar to the ones used in SLAM applications, high accuracy and real-time performance is yet to be achieved.
Furthermore, these approaches require a large number of contacts, which may be impractical for effective object manipulation.
Nevertheless, it is foreseen that tactile sensors of different types will need to be used along with vision for solving the localization problem.
Due to the reduced number of contacts, high resolution tactile sensors that can capture the object's local shape can be useful to determine the pose using few contacts.


\section{Tactile sensing in sensor fusion}\label{sensingintegration}

Robots must be equipped with different sensing modalities to be able to operate in unstructured environments.
The fusion of these different sources of data into more meaningful, higher-level representations of the state of the world is also part of the process of perception (See Fig. \ref{fig:tactilesensingsystem}).
The information that is provided by the sensors may be redundant, reducing the uncertainty with which features are perceived by the system, and increasing the reliability of the system in case of failure \cite{felip2014multisensor}.
Combining different sensing modalities which provide complementary information may also have a synergistic effect, where features in the environment can be perceived in situations where it would be impossible using the information from each sensor separately \cite{luo2002multisensor}.
Furthermore, multiple sensors can provide more timely and less costly information, given the different operating speeds of the sensors and the fact that their information can be processed in parallel.
\\
In tasks that require interaction with the environment, tactile sensing can be combined with other sensing modalities to increase precision and robustness  \cite{prats2009vision}. 
Most typical sensing arrangements in robot manipulation and other tasks that include physical interaction, are combinations of tactile sensing with vision, kinaesthetic cues, force-torque and range sensing \cite{luo1989multisensor}. We have discussed the works on integrating tactile sensing and kinaesthetic cues for global shape recognition in Section~\ref{globalshaperecognition}. In this section, we mainly focus on tactile sensing in sensor fusion with vision.

In robotics, attempts to fuse vision and touch to recognize and represent objects can be dated back to the 1980's \cite{allen1984surface}. In most applications, tactile sensing was utilized to support vision for improving performance in object recognition \cite{bjorkman2013enhancing}, object reconstruction \cite{ilonen2013fusing} and grasping \cite{bekiroglu2013probabilistic}. With the increasing performance in the last few decades, tactile sensors have shown the potential to play a more significant role in tasks using information integrated from different modalities \cite{sinapov2011interactive,hebert2011fusion,araki2012online}.
Thanks to the development of tactile sensor technologies, the role of tactile sensing in sensor fusion for multiple applications has evolved to a mature stage in a number of applications, which can be summarized as follows:

\paragraph*{Verifying contacts} Most early researchers took tactile sensors as devices to verify contacts due to their low resolution. In this type of methods, rough object models are first built by vision and the description is then refined and detailed by tactile sensing. For instance, in \cite{allen1984surface} vision is first used to obtain object contours and edges as it can capture rich information rapidly, by taking bi-cubic surface patches as primitives. The tactile trace information is then added into the boundary curves acquired in the first step to achieve a more detailed description of the surfaces. In this work, tactile (haptic) sensing is utilized to trace the object surface to get 3D coordinates, surface normal and tangential information of contact points. A similar framework is also employed in \cite{allen1988integrating}, where the role of tactile sensing is greater. The strategy consisted of exploring regions that are uncertain for vision, determining the features such as surfaces, holes or cavities, and outlining the boundary curves of these features. In \cite{allen1999integration} vision is used to estimate the contact positions along a finger and applied forces for grasping tasks, whereas tactile sensors and internal strain gauges are utilized to assist vision, especially in two cases: when vision is occluded or when it is needed to further confirm contact positions determined by vision.

In \cite{dragiev2011gaussian}, implicit surface potentials described by Gaussian Processes are taken as object models for internal shape representations. In the Gaussian Process of shape estimation, uncertain sensor channels, i.e., tactile, visual and laser, are equally integrated, to support the control of reach-and-grasp movements. In \cite{ilonen2013fusing}, an optimal estimation method is proposed to learn object models during grasping via visual and tactile data by using iterative extended Kalman filter (EKF). A similar work is carried out in \cite{bjorkman2013enhancing} but Gaussian process regression is utilized instead of EKF: visual features are extracted first to form an initial hypothesis of object shapes; tactile measurements are then added to refine the object model. Though tactile array sensors are employed in these works, still only the locations of tactile elements that are in contact with the object are used, not including the information of pressure distribution in tactile arrays. A framework to detect and localize contacts with the environment using different sources of sensing information is presented in \cite{felip2014multisensor}. Each sensor generates contact hypotheses which are fused together using a probabilistic approach to obtain a map that contains the likelihood of contact at each location in the environment.
This approach was tested in different platforms with various sensing modalities, such as tactile, force-torque, vision, and range sensing.

\paragraph*{Extracting features to assist vision} In \cite{bekiroglu2013probabilistic}, multiple sensory streams, i.e., vision, proprioception and tactile sensing, are integrated to facilitate grasping in a goal-oriented manner. In this work, image moments of tactile data (both 2D and 3D), are utilized; together with vision and action features, a feature set is formed for grasping tasks. In \cite{hebert2011fusion}, the in-hand object location is estimated by fusing data from stereo vision, force-torque and joint sensors. The sensor fusion is achieved by simply concatenating features or data from single modalities into a joint vector. In \cite{guler2014s}, vision and tactile sensing are proved to be complementary in the task of identifying the content in a container by grasping: as the container is squeezed by a robot hand, deformation of the container is observed by vision and the pressure distributions around the contact region are captured by tactile sensors. PCA is employed to extract PCs from vision, tactile and vision-tactile data, which are then fed into classifiers. It is concluded that combining vision and tactile sensing leads to the improvement of classification accuracy where one of them is weaker. In addition, some researchers have attempted to combine vision, tactile and force sensing to estimate the object pose in \cite{bimbo2015global}.

\paragraph*{Providing local and detailed information} In \cite{schmid2008opening}, a multi-sensor control framework is created for a task of opening a door. The vision is used to detect the door handle prior to the gripper-handle contact and tactile sensors mounted on the robot grippers are used to measure the gripper orientation with respect to the door handle by applying moment analysis to tactile readings during the gripper-handle contact. Based on the information, the pose of grippers can be adjusted accordingly. In \cite{prats2010reliable}, vision, force/torque sensors and tactile sensors are also combined in a control scheme for opening a door. A sensor hierarchy is established in which the tactile sensing is preferred over vision: the tactile sensing can obtain robust and detailed information about the object position, whereas vision provides global but less accurate data. In both works, tactile sensors play a greater role than just confirming contacts but the vision and tactile sensing are employed in different phases instead of achieving a synthesized perception. In recent work \cite{izatt2017tracking}, a GelSight sensor is taken as a source of dense local geometric information that is incorporated in object point cloud from an RGB-D camera.

\paragraph*{Transferring knowledge with vision} The information of object representations has been found to be shared by vision and tactile sensing of humans \cite{lacey2007vision} and the visual imagery is discovered to be involved in the tactile discrimination of orientation in normally sighted humans \cite{zangaladze1999involvement}. It has also been shown that the human brain employs shared models of objects across multiple sensory modalities such as vision and tactile sensing so that knowledge can be transferred from one to another \cite{amedi2002convergence}. This mechanism can also be applied in robotic applications. In \cite{kroemer2011learning}, vision and tactile samples are paired to classify materials using a dynamic tactile sensor. In the training phase, material representations are learned by both tactile (haptic) and visual observations of object surfaces. A mapping matrix for transferring knowledge from vision to tactile domain is then learned by dimensionality reduction and at the test stage, materials can be classified with only tactile (haptic) information available based on the obtained shared models. In \cite{sanchez2013sensorimotor}, vision and tactile feedback are associated by mapping visual and tactile features to object curvature classes. By taking advantage of the fact that both visual and tactile images are present in numerical matrices, a tactile sensor can be localized in a visual object map by sharing the same sets of features between the visual map and tactile readings \cite{luo2015localizing}. In recent works \cite{gao2016deep} and \cite{yuan2017connecting}, deep networks, typically Convolutional Neural Networks (CNNs), are applied to both vision and hatpic data and the learned features from both modalities are then associated by a fusion layer for haptic classification.


To conclude, as the performance of tactile sensors improves, tactile sensing plays an increasingly important role in the sensor fusion for multiple applications. In addition to providing information of contact locations like single-point contact sensors, more useful cues can be extracted from tactile readings, such as detailed local information and features to improve models built by vision. More importantly, knowledge can be transferred between vision and tactile sensing, which can improve the perception of the environment interacting with robots.

\section{Discussion and conclusion}\label{discussionandconclusion}
The rapid development of tactile devices and skins over the last couple of years has provided new opportunities for applying tactile sensing in various robotic applications. Not only the types of tactile sensors are to be considered for an application, the suitable computational method to decipher the encoded information is also of the utmost importance. At the same time it has brought to the fore numerous challenges towards effective use of tactile data. This paper focuses on tactile object recognition of shape and surface material and the estimation of its pose. Knowledge of these properties is crucial for the successful application of robots in grasping and manipulation tasks. In this section the progress in the literature is summarised, open issues are discussed and future directions are highlighted.

\paragraph*{Selection of tactile sensors to meet the requirements of tactile object recognition and localization} Apart from the algorithms used to interpret sensor data, the recognition performance also depends on what tactile sensors are used. As high spatial and magnitude resolution provides more detailed information on the object, tactile sensors of higher resolution are preferable. But on the other hand, higher resolution will likely bring about higher costs, both in development and fabrication of the sensor. Most of the available tactile sensors used in robotic fingertips are of lower or similar resolution over an equivalent area compared to the spatial resolution of the human finger, i.e., the density of Merkel receptors in the fingertip (14$\times$14) \cite{pezzementi2011tactile}. For instance, the widely used tactile sensor modules in the literature Weiss DSA 9205 (14$\times$6) with a size of 24 mm by 51 mm; the Weiss DSA 9335 sensor module provides a similar spatial resolution of 3.8 mm with 28 sensor cells organized as 4$\times$7 matrices; the TakkTile array sensor senses independent forces over an 8$\times$5 grid with approximately 6 mm resolution \cite{tenzer2014feel}; the DigiTacts sensor from Pressure Profile Systems has 6$\times$6 sensor elements with 6$\times$6 mm resolution; a Tekscan sensor (model: 4256E) has a sensing pad of 24 mm by 35 mm with 9$\times$5 elements. To achieve good spatial resolution, these sensors are commonly costly. An exception is the GelSight tactile sensor that has high resolution of 320$\times$240 and is economical, as it utilizes a low-cost webcam to capture the object properties. But the inclusion of a webcam and illumination LEDs makes the whole sensor bulky and difficult to miniaturize. Therefore, researchers need to balance multiple factors when selecting tactile sensors, i.e., spatial and magnitude resolution, size and cost of the sensors.

In real-time tactile localization applications, it is required that tactile sensors have quick responses and short latency. Human tactile receptors can sense the transient properties of a stimulus at up to 400 Hz \cite{jamali2011majority}. Since the output of single-point contact sensors, i.e., force sensors and whisker based sensors, is of low dimension, these sensors have typically high update rate and low latency. For instance, the data-acquisition of strain gauges and PVDFs embedded in a robotic finger can achieve a temporal resolution of 2.5 kHz \cite{jamali2011majority}. In contrast, tactile array sensors output data of higher resolution and therefore need to reduce the latency and improve response times. The maximum scanning rate for the Weiss DSA 9205 sensor is around 240 frames per second (fps) \cite{schneider2009object}, while using a serial bus (RS-232) shared by multiple sensors for data transmission will limit the available frame rate for each sensor to approximate 30 Hz when six sensor pads are used \cite{meier2011probabilistic}; the update rate of the Tekscan pressure sensing is set as 200 Hz in \cite{liu2012tactile}; by using a GelSight sensor, the in-hand localization can be achieved at 10 fps \cite{li2014localization}. An alternative method to sense the transient stimulus is to use neuromorphic tactile systems that provides transient event information as the sensor interface with objects \cite{lee2015kilohertz,bartolozzi2016neuromorphic}. The sensing element triggers an event when it detects relevant transient stimulus and this provides important local information in manipulation \cite{son1994tactile}.

Many of the tactile sensors used in the literature are of rigid sensing pad and plated with an elastic cover such as Weiss modules and TakkTile sensors. As the data are read from sensing elements arranged in a plane, it reduces the complexity of data interpretation but these sensors are hard to be mounted on curved robot body parts. There are also flexible tactile sensors like DigiTacts from PPS and Tekscan tactile system. They can be attached to curved surfaces but the data interpretation becomes more complex.


\paragraph*{Tactile object recognition} As for the local shape recognition by touch sensing, multiple feature descriptors have been proposed in the literature to represent features within tactile readings. The ``do-nothing" descriptors, statistical features, and PCA-based features are simple and easy to implement but normally sensitive to pattern variances. Vision-based descriptors have been observed to be the mainstream by taking tactile readings as images. It can be foreseen that more vision descriptors will be applied in touch sensing to represent the information embedded in the tactile data which may provide cues of how to combine vision and touch sensing. In addition, unsupervised learning and deep architectures show promise in recent years by obtaining self-organizing features from raw tactile readings without pre-defining the feature structure. To seek shape representation appropriate for tactile sensing, these methods originated from vision could be further investigated by examining the differences between vision and tactile sensing mechanisms.

To reveal the global object shapes, one method is to utilize the distribution of contact points that can retrieve arbitrary contact shapes but will be time consuming when investigating a large object surface, while also being sensitive to object movements. The other method is to use local tactile patterns and a paradigm of this method is Bag-of-Features that is becoming popular recently. It can reveal local features but the three-dimensional distribution information of the object is not incorporated. One future direction is to perceive global shapes by integrating tactile patterns and kinaesthetic cues to achieve a better recognition performance as cutaneous sensing and proprioception are correlated in object recognition.

\paragraph*{Object localization by touch sensing} The object localization via touch could be a significant application that can facilitate object manipulation and can be complementary to visual localization. Currently, most of the attention regarding this topic has been paid to haptic-based localization with single-point contact sensors. But single-point contact sensors, like force sensors and whiskers, provide only limited information, i.e., contact locations of single points. Single-point contact sensors also require multiple contacts, which may cause the unpredicted motion of the object during the contacts. Thanks to the advancement of tactile devices, the works into other types of localization have emerged, especially using tactile array sensors to identify contact patterns. There exist multiple open issues expected to be investigated: 1) how to map the contact points on the sensor pad to the 3D space; 2) how to combine the data points acquired from vision and tactile sensing; 3) how to define the priority hierarchy when the information from different sensors conflicts.

\paragraph*{Tactile sensing in sensor fusion} The use of tactile sensing has also been applied to material recognition and some other emerging applications such as slip detection, grasp stability assessment, identification of touch modalities and tactile servoing. The role of tactile sensing plays in sensor fusion has evolved from just verifying object-sensor contacts to extracting features to assist vision and provide detailed local information, and to transferring knowledge with vision. It is envisioned that tactile sensing will play a more important role in multiple applications. The development of a ``sensor hierarchy" could be sought to establish the ``preference" of sensor data. 

\paragraph*{Deep learning in robotic tactile perception} After dominating the computer vision field, deep learning, especially CNNs, has also been applied in robotic tactile perception in recent years, from object shape recognition \cite{schmitz2014tactile,cao2016efficient}, hardness estimation \cite{yuan2017shape}, to sharing features with vision \cite{gao2016deep,yuan2017connecting}. When using deep learning, there are several issues need to be carefully taken care of. 1) \textbf{\textit{Why deep learning?}} Deep learning can learn self-organising features instead of defining hand-crafted features a-priori and has shown considerable success in tasks such as object recognition. However, online application of such computationally intensive algorithm is difficult, especially for tasks need timely response such as grasping and manipulation, and it is hard to tune the parameters of deep architectures. On the other hand, it would be an overkill to put deep learning on small datasets and simple tasks; such algorithms can also potentially lose physical interpretation. 2) \textbf{\textit{Data collection.}} Compared to datasets in vision, it is relatively expensive to collect tactile data, not only because of the required robot components but also the efforts that go into designing the data collection process. On the other hand, the same algorithm may achieve distinct performances on data collected from different tactile sensors. Due to such reasons, there are still no commonly used datasets for evaluating the processing algorithms in tactile sensing but a substantially large haptic dataset would benefit the field. 3) \textbf{\textit{Labeling and ground truth.}} It is hard to obtain ground truth for tactile perception tasks using crowd-sourcing tools like Amazon Mechanical Turk in vision. In the current literature \cite{gao2016deep,yuan2017connecting}, tactile properties such as haptic adjectives (e.g., compressible or smooth) are labeled by limited number of human annotators which may bring bias to the ground truths. To this end, domain adaptation can be applied from vision to tactile sensing and unsupervised learning will be of benefit in the future research of tactile sensing.

\section*{Acknowledgment}
The work presented in this paper was partially supported by the Engineering and Physical Sciences Council (EPSRC) Grant (Ref: EP/N020421/1), the King's-China
Scholarship Council PhD scholarship, the European Commission under grant agreements PITN–GA–2012–317488-CONTEST, and EPSRC Engineering Fellowship for Growth – Printable Tactile Skin (EP/M002527/1).




%
\bibliographystyle{IEEEtran}
\bibliography{main} 

%

\begin{IEEEbiography}[{\includegraphics[width=1in,height=1.25in,clip,keepaspectratio]{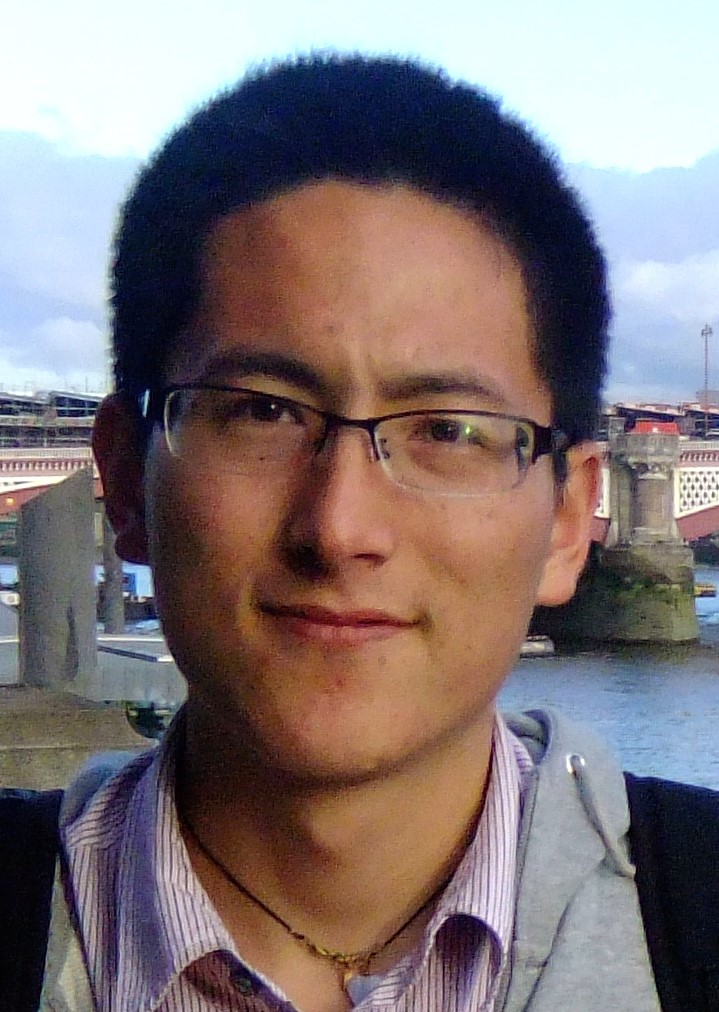}}]{Shan Luo}
is currently a Postdoctoral Research Fellow at Brigham and Women's Hospital, Harvard Medical School. Previous to this position, he was a Research Fellow at University of Leeds and a Visiting Scientist at Computer Science and Artificial Intelligence Laboratory (CSAIL), Massachusetts Institute of Technology (MIT). He received the B.Eng. degree in Automatic Control from China University of Petroleum, Qingdao, China, in 2012. He was awarded the Ph.D. degree in Robotics from King's College London, UK, in 2016. His research interests include tactile sensing, object recognition and computer vision.
\end{IEEEbiography}

\vspace{2 cm}
\begin{IEEEbiography}[{\includegraphics[width=1in,height=1.25in,clip,keepaspectratio]{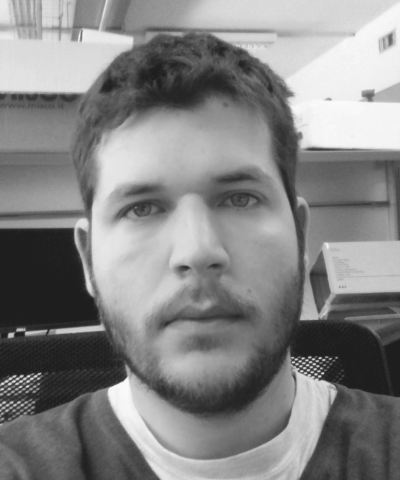}}]{Joao Bimbo}
is currently a Postdoctoral Researcher at the Istituto Italiano di Tecnologia. He received his MSc. degree in Electrical Engineering from the University of Coimbra, Portugal, in 2011. He was awarded the Ph.D. degree in Robotics from King's College London, UK, in 2016. His research interests include tactile sensing and teleoperation.
\end{IEEEbiography}

\vspace{5 cm}
\begin{IEEEbiography}[{\includegraphics[width=1in,height=1.25in,clip,keepaspectratio]{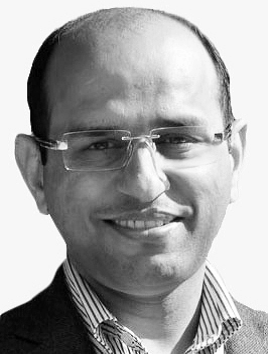}}]{Ravinder Dahiya}
is Professor and EPSRC Fellow in the School of Engineering at University of Glasgow, UK. He is Director of Electronics Systems Design Centre in University of Glasgow and leader of Bendable Electronics and Sensing Technologies (BEST) group. He completed Ph.D. at Italian Institute of Technology, Genoa (Italy). In past, he worked at Delhi University (India), Italian Institute of Technology, Genoa (Italy), Fondazione Bruno Kessler, Trento (Italy), and University of Cambridge (UK).

His multidisciplinary research interests include Flexible and Printable Electronics, Electronic Skin, Tactile Sensing, robotics, and wearable electronics. He has published more than 160 research articles, 4 books (including 3 in various stage of publication) and 9 patents (including submitted). He has worked on and led many international projects.

He is Distinguished Lecturer of IEEE Sensors Council and senior member of IEEE. Currently he is serving on the Editorial Boards of Scientific Reports (Nature Pub. Group), IEEE Transactions on Robotics and IEEE Sensors Journal and has been guest editor of 4 Special Journal Issues. He is member of the AdCom of IEEE Sensors Council and is the founding chair of the IEEE UKRI sensors council chapter. He was General Chair of IEEE PRIME 2015 and is the Technical Program Chair (TPC) of IEEE Sensors 2017.

Prof. Dahiya holds prestigious EPSRC Fellowship and also received Marie Curie Fellowship and Japanese Monbusho Fellowship. He was awarded with the University Gold Medal and received best paper awards 2 times and another 2 second best paper awards (as co-author) in the IEEE international conferences. He has received numerous awards including the 2016 IEEE Sensors Council Technical Achievement award and 2016 Microelectronics Young Investigator Award. Dr. Dahiya was among Scottish 40under40 in 2016.
\end{IEEEbiography}

\vspace{-1 cm}
\begin{IEEEbiography}[{\includegraphics[width=1in,height=1.25in,clip,keepaspectratio]{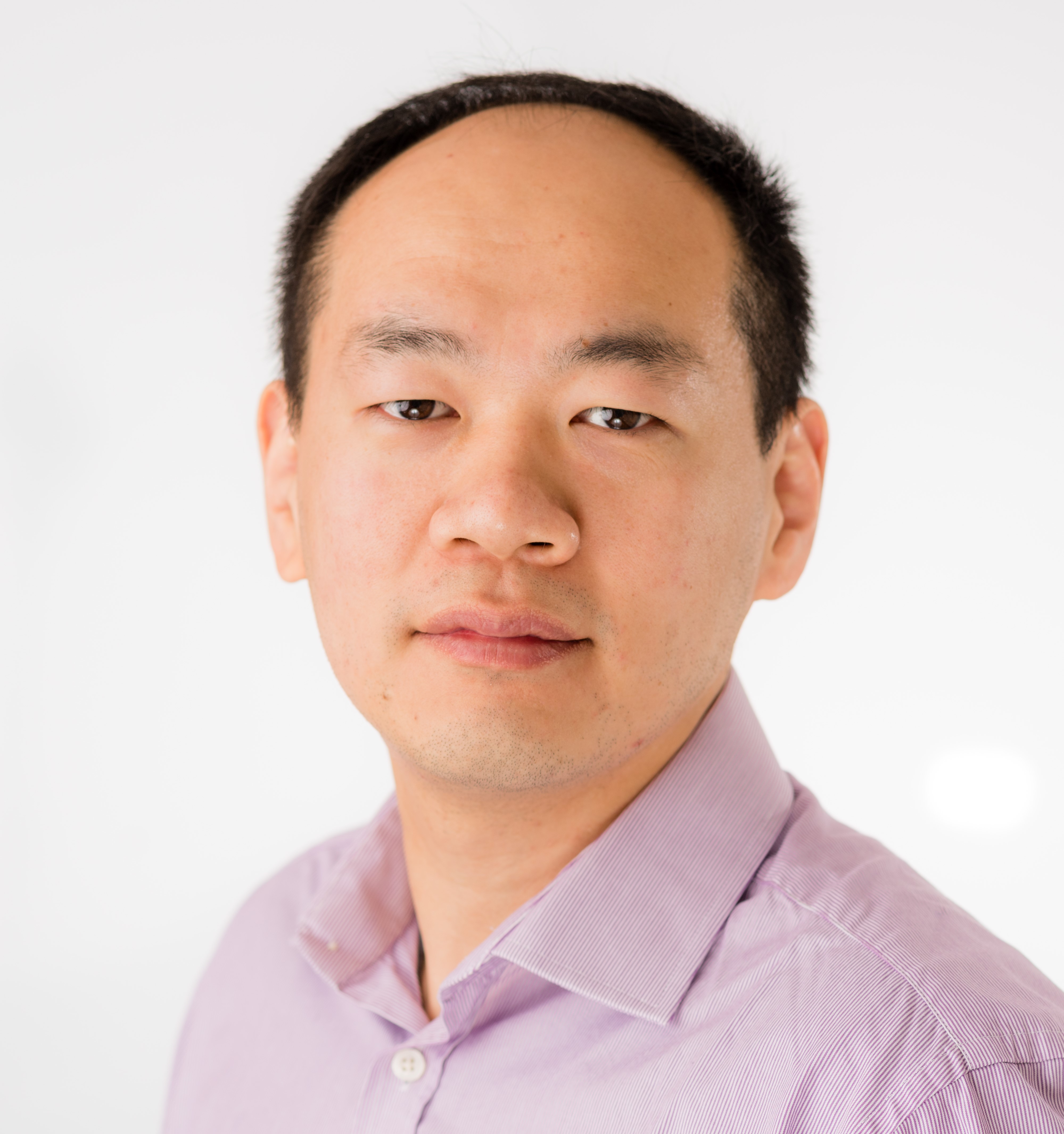}}]{Hongbin Liu}
is a Senior Lecturer (Associate Professor) in the Department of Informatics, King's College London (KCL) where he is directing the Haptic Mechatronics and Medical Robotics Laboratory (HaMMeR) within the Centre for Robotics Research (CoRe). Dr. Liu obtained his BEng in 2005 from Northwestern Polytechnical University, China, MSc and PhD in 2006 and 2010 respectively, both from KCL. He is a Technical Committee Member of IEEE EMBS BioRobotics. He has published over 100 peer-reviewed publications at top international robotic journals and conferences. His research lies in creating the artificial haptic perception for robots with soft and compliant structures, and making use of haptic sensation to enable the robot to effectively physically interact with complex and changing environment. His research has been funded by EPSRC, Innovate UK, NHS Trust and EU Commissions.
\end{IEEEbiography}



\flushend
\end{document}